\documentclass[letterpaper]{article} 
\usepackage{aaai2026}  
\usepackage{times}  
\usepackage{helvet}  
\usepackage{courier}  
\usepackage[hyphens]{url}  
\usepackage{graphicx} 
\usepackage{natbib}  
\usepackage{caption} 
\usepackage{algorithm}
\usepackage{amsmath}
\usepackage{xcolor}
\usepackage{algpseudocode}
\usepackage{multirow}
\algrenewcommand\alglinenumber[1]{\footnotesize #1:}

\usepackage{newfloat}
\usepackage{listings}
\DeclareCaptionStyle{ruled}{labelfont=normalfont,labelsep=colon,strut=off} 
\floatstyle{ruled}
\newfloat{listing}{tb}{lst}{}
\floatname{listing}{Listing}
\title{SCOUT-RAG: Scalable and Cost-Efficient Unifying Traversal for\\ Agentic Graph-RAG over Distributed Domains}

\author{
    Longkun Li\textsuperscript{\rm 1}\thanks{Work done during an internship at A*STAR.},
    Yuanben Zou\textsuperscript{\rm 2,*},
    Jinghan Wu\textsuperscript{\rm 3,*},
    Yuqing Wen\textsuperscript{\rm 4,*}\\
    Jing Li\textsuperscript{\rm 5, \rm 6}\thanks{Corresponding Author}
    Hangwei Qian\textsuperscript{\rm 5,\rm 6},
    Ivor Tsang\textsuperscript{\rm 5, \rm 6}
}

\affiliations{
    \textsuperscript{\rm 1}Yong Loo Lin School of Medicine, National University of Singapore, Singapore\\
    \textsuperscript{\rm 2}College of Design and Engineering, National University of Singapore, Singapore\\
    \textsuperscript{\rm 3}Institute of Artificial Intelligence and Robotics, Xi'an Jiaotong University, China\\
    \textsuperscript{\rm 4}Faculty of Science, National University of Singapore, Singapore\\
    \textsuperscript{\rm 5}Institute of High Performance Computing, Agency for Science, Technology and Research, Singapore\\
    \textsuperscript{\rm 6}Centre for Frontier AI Research, Agency for Science, Technology and Research, Singapore\\
    \{e1324260, e1142979, e1351400\}@u.nus.edu, wujing4022@stu.xjtu.edu.cn,\\
    \{li\_jing, qian\_hangwei, ivor\_tsang\}@a-star.edu.sg
}

\usepackage{bibentry}

\begin{document}

\maketitle

\begin{abstract}
Graph-RAG improves LLM reasoning using structured knowledge, yet conventional designs rely on a centralized knowledge graph. In distributed and access-restricted settings (e.g., hospitals or multinational organizations), retrieval must select relevant domains and appropriate traversal depth without global graph visibility or exhaustive querying. 
To address this challenge, we introduce \textbf{SCOUT-RAG} (\textit{\underline{S}calable and \underline{CO}st-efficient \underline{U}nifying \underline{T}raversal}), a distributed agentic Graph-RAG framework that performs progressive cross-domain retrieval guided by incremental utility goals. SCOUT-RAG employs four cooperative agents that: (i) estimate domain relevance, (ii) decide when to expand retrieval to additional domains, (iii) adapt traversal depth to avoid unnecessary graph exploration, and (iv) synthesize the high-quality answers. The framework is designed to minimize retrieval regret, defined as missing useful domain information, while controlling latency and API cost. Across multi-domain knowledge settings, SCOUT-RAG achieves performance comparable to centralized baselines, including DRIFT and exhaustive domain traversal, while substantially reducing cross-domain calls, total tokens processed, and latency.
\end{abstract}


\section{Introduction}

RAG (Retrieval-Augmented Generation)~\cite{lewis2020retrieval} enhances large language models by retrieving and integrating external knowledge for grounded response generation. Traditional deployments primarily rely on vector-based retrieval, where individually relevant text segments are retrieved based on similarity signals. Yet many real-world queries demand global sense-making, synthesizing relationships, dependencies, and abstractions across entities rather than isolating local matches. To address this gap, Graph-RAG systems~\cite{edge2024local} have emerged, enabling multi-hop traversal, hierarchical abstraction, and structured knowledge integration. These capabilities have led major industry platforms, including Microsoft, NebulaGraph, AntGroup, and Neo4j~\cite{peng2024graph}, to adopt graph-centric retrieval pipelines, underscoring the emerging role of structured knowledge access in complex information environments.

However, most RAG and Graph-RAG frameworks assume centralized access to a unified knowledge store. In practice, data often resides across independent organizations that cannot share raw content due to privacy, ownership, or regulatory constraints, motivating retrieval across distributed and siloed domains rather than a single graph. This setting has been formalized as multi-domain or federated RAG~\cite{shojaee2025federated}, where each domain retains data control and responses must respect access boundaries. Further, each cross-domain request may incur latency or monetary costs, requiring systems that minimize unnecessary retrieval while preserving answer fidelity. Recent work uses supervised domain routers~\cite{shojaee2025federated,guerraoui2025efficient} to filter domains, but these approaches depend on labeled query–domain pairs and do not adapt to evolving utility signals or cold-start deployments.

Distributed Graph-RAG features partial observability, heterogeneous access constraints, and incremental evidence accumulation, which collectively motivate an \textbf{agentic} formulation. Rather than predetermining a fixed retrieval plan, the system must treat cross-domain traversal as a sequential decision process: evaluating intermediate signals, updating beliefs about domain relevance, and allocating retrieval budget adaptively across local exploration and global expansion. This perspective naturally accommodates privacy constraints, variable query structure, and shifting domain utility without requiring labeled access patterns or centralized visibility. By framing retrieval as iterative utility estimation and constrained action selection, an agentic approach supports controlled cost–quality trade-offs and scalable reasoning across distributed knowledge holders.

Building on this perspective, we present \textbf{SCOUT-RAG}, a \textbf{layered agentic architecture for distributed Graph-RAG}, explicitly designed to operate under cost, privacy, and scalability constraints. Our design is grounded in three core insights. \textbf{First}, we introduce training-free domain relevance estimation, avoiding supervised routing signals and enabling deployment in cold-start and privacy-restricted settings. \textbf{Second}, we generalize Graph-RAG to decentralized knowledge holders, treating each domain as a partially observable subgraph and enabling adaptive switching between local traversal and cross-domain expansion as context evolves. \textbf{Third}, motivated by agent-based RAG systems~\cite{singh2025agentic} and recent agentic Graph-RAG efforts~\cite{leeagent,Graph-R1}, we formalize retrieval as a sequential decision process that incrementally assesses utility, allocates exploration budget, and adjusts traversal depth and breadth under explicit cost constraints. Together, these principles enable scalable, privacy-aware, and cost-efficient cross-domain knowledge aggregation without assuming centralized visibility or requiring supervised domain routing. 
To the best of our knowledge, \textbf{SCOUT-RAG is the first to operationalize scalable agentic Graph-RAG across independent knowledge domains}, supporting adaptive multi-domain traversal under explicit cost and privacy objectives.

Overall, the main contributions of this work are as follows:
\begin{itemize}
    \item We introduce the first cross-domain Graph-RAG setting, where knowledge is distributed across independent data holders and cross-domain access is gated by privacy and cost constraints. This setting highlights the scalability and utility challenges of decentralized sense-making queries.
    \item We propose SCOUT-RAG, a layered agentic system that conducts training-free domain relevance estimation, local vs.~global traversal decisions, and adaptive depth–breadth exploration across multiple domains.
    \item We design a closed-loop evaluation and improvement mechanism in which assessment agents monitor answer sufficiency and factual coverage, guiding strategy agents to selectively expand domains or deepen graph traversal when needed.
    \item SCOUT-RAG implements multi-level safeguards, including strict time-budget enforcement, best-answer tracking, and coordinated parallelism, to ensure reliable execution and prevent cascading failures in decentralized access settings.
    \item Evaluations on 89 multi-domain queries across 1–40 countries show that SCOUT-RAG outperforms centralized local and global baselines, while nearly matching centralized DRIFT (56 vs. 63) with over 4× fewer tokens and lower latency, demonstrating efficient and scalable decentralized retrieval.
\end{itemize}

\section{Related Work}

\textbf{Gragh Retrieval-Augmented Generation.} RAG improves language model grounding by retrieving external evidence~\cite{lewis2020retrieval}. GraphRAG further enhances contextual reasoning by leveraging graph structure for hierarchical retrieval and local exploration~\cite{edge2024local}. Typically, users are required to specify the retrieval mode—global or local—when using Graph-RAG systems, in order to indicate which level of granularity is preferred for the current query. Recent advances have explored organizing retrieved information through hierarchical summary trees~\cite{sarthi2024raptor}, constructing entity graphs from document-level chunks~\cite{zhao20252graphrag}, or extracting entities and relations with industrial-grade NLP libraries~\cite{min2025efficient}. However, these methods work under a unified corpus or global knowledge graph, which can be complementary for the problem setup where information is siloed, access is constrained, and retrieval must remain selective and budget-aware. The DRIFT framework~\footnote{\url{https://microsoft.github.io/graphrag/query/drift_search/}} from Microsoft Research introduced the idea of combining global reasoning with targeted local inspection to improve retrieval efficiency. Our proposed framework draws inspiration from this design: SCOUT-RAG adopts DRIFT-style global–local balancing but extends it to a multi-domain setting, where retrieval policies must reinterpret the notions of "global" and "local" in order to determine the most appropriate level of granularity for each specific query.

\textbf{Cross-Domain Traversal.}
Beyond domain-specific reasoning in Graph-RAG~\cite{xiao2025graphrag}, recent practical applications have highlighted the need to retrieve information across multiple domains~\cite{liu2024raemollm,shojaee2025federated}, which calls for more principled search strategies. Ideally, domains can be accessed independently and concurrently, allowing overall latency to be reduced as multiple domain searches proceed in parallel. However, in practice, latency may be amplified due to bandwidth and network overhead, the dominance of the slowest domain, or additional computation and synchronization delays across domains. A recent automated platform for multi-domain evaluation of RAG systems, OmniBench-RAG~\cite{liang2025omnibench}, has incorporated efficiency as a key dimension in performance quantification. In addition, many systems, such as KG-GPT~\cite{kim2023kg} and StructGPT~\cite{jiang2023structgpt}, have integrated LLM-based retrievers, implying that exhaustive retrieval across all domains would be costly. Regarding the efficiency problems, existing approaches rely on labeled data to train models that identify relevant domains, commonly referred to as domain routers~\cite{shojaee2025federated,guerraoui2025efficient}. Our proposed SCOUT-RAG leverages an agent to rank domain relevance and dynamically explore appropriate domains based on multiple key signals. The agent can operate under a cold-start condition and becomes increasingly reliable as more queries are accumulated. To enhance usability, we also introduce explicit termination conditions, such as time tolerance and answer-quality constraints.

\textbf{Agentic AI Systems.} Recent advances in multi-agent systems explore coordinated LLM roles for complex reasoning and tool-use workflows. AutoGen~\cite{wu2024autogen} demonstrates conversational collaboration among agents, while MetaGPT~\cite{hong2023metagpt} formalizes structured role assignments inspired by software engineering teams. ReAct~\cite{yao2022react} couples reasoning traces with action execution for controllability. Several pioneering works have explored the integration of agentic AI into Graph-RAG systems. KG-R1~\cite{song2025efficient} employs a single agent that interacts with the knowledge graph as its environment, learning to retrieve relevant information step by step and integrating it into its reasoning and text generation process. Agent-G~\cite{leeagent} introduces feedback loops that integrate an agent to improve retrieval quality, where an auxiliary critic module rejects unsatisfactory answers. We build on these insights but shift focus to governing retrieval decisions for multi-sources scenarios: SCOUT-RAG operationalizes four specialized agents for domain assessment, evidence extraction, synthesis, and iterative refinement, each explicitly grounded in cross-domain information access needs.

\section{Methodology}

\subsection{Problem Formulation}

We consider retrieval-augmented generation over $M$ distributed knowledge domains. 
Each domain $i \in \{1,\dots,M\}$ maintains a private knowledge graph 
$\mathcal{G}_i = (\mathcal{V}_i, \mathcal{E}_i, \mathbf{X}_i)$ consisting of entities, 
relations, and associated text or structured attributes. 
Raw data cannot be shared across domains due to privacy and ownership constraints.  
Given a natural-language query $q$, the system aims to synthesize an answer $A$ grounded in evidence retrieved from multiple domains, without centralizing the underlying graphs and adhering to a prescribed cost budget.

Let $\pi_i$ denote the retrieval policy for domain $i$, specifying whether and how to perform retrieval, with associated cost $c_i$ (e.g., time or computational resources).  
The objective is to produce an answer that maximizes the answer quality:
\begin{equation}
\begin{aligned}
& \max_{A} \quad 
    \mathrm{Qual}(A \mid q, \,\, \{\pi_i\}_{i=1}^M) \\
\text{s.t.} \quad 
    & A = \Phi\!\big(\{f_{\pi_i}\!\big(q, \mathcal{G}_i\big)\}_{i=1}^M\big), \quad  \sum_{i=1}^{M} c_i \le \mathcal{C}_{\max},
\end{aligned}
\end{equation}
where $f_{\pi_i}(\cdot)$ is a generator producing a local answer $A_i$ under policy $\pi_i$, $\Phi(\cdot)$ denotes the aggregation function synthesizing the final answer across domains, 
and $\mathrm{Qual}(\cdot)$ quantifies the grounding fidelity and relevance of the generated answer with respect to retrieved evidence.

In practice, the retrieval policy $\{\pi_i\}$ is not executed in a single pass. 
Instead, \textsc{SCOUT-RAG} operates through \emph{progressive policy refinement}:  
(i) initial estimation of domain relevance, 
(ii) domain-scoped partial retrieval and answer seeding, and 
(iii) iterative improvement via local depth or breadth expansions guided by answer-quality 
feedback and remaining budget. This sequential decision process incrementally allocates 
retrieval resources while honoring $\mathcal{C}_{\max}$, terminating when no further utility is gained 
or the budget is exhausted.

\begin{figure}[t]
    \centering
    \includegraphics[width=\columnwidth]{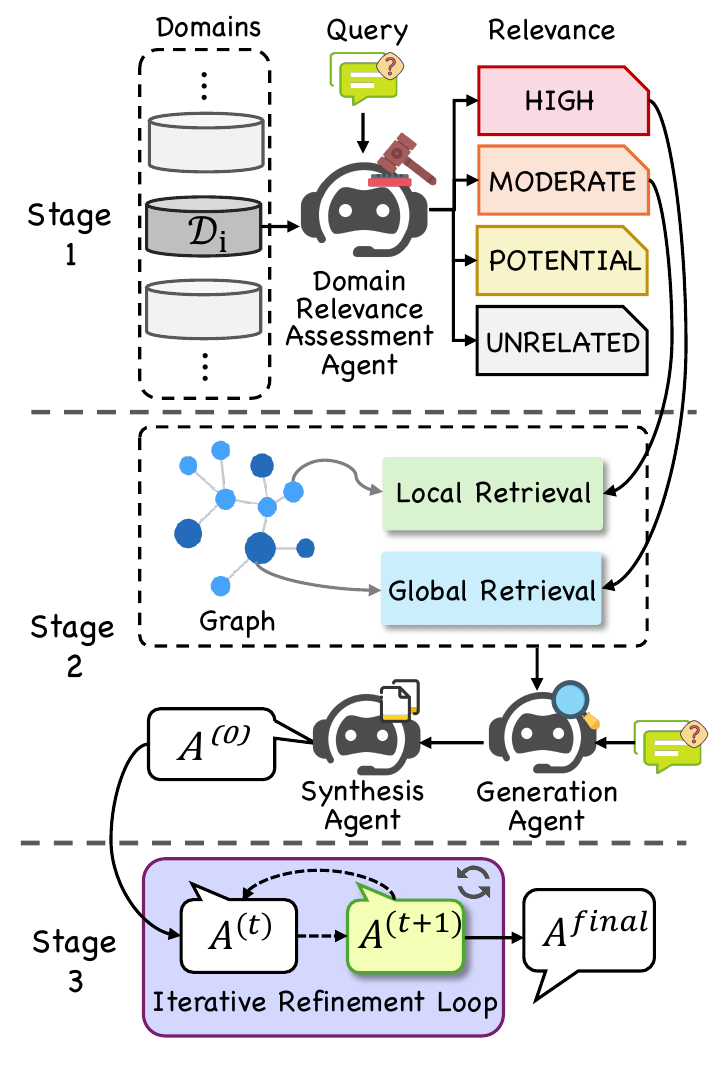}
    \caption{
    Overview of the proposed \textbf{SCOUT-RAG} framework. 
    The framework operates in three stages: (i) domain relevance estimation, (ii) domain-aware retrieval (local vs. global), and (iii) iterative answer refinement via synthesis and generation agents. These agentic components collectively support adaptive, cross-domain reasoning without centralized graph access. 
    }
    \label{fig:pipeline}
\end{figure}

This formulation captures the fundamental characteristics of distributed Graph-RAG: 
(i) \emph{partial observability}, since the global graph $\bigcup_{i=1}^{M} \mathcal{G}_i$ is never exposed; 
(ii) \emph{adaptive cross-domain reasoning}, requiring decisions on when to remain local vs.\ 
expand to new domains; and 
(iii) \emph{cost-aware retrieval refinement}, where utility estimates guide incremental evidence gathering rather than exhaustive querying.
Building upon this formulation, we design a decentralized solution that operates the above objectives through coordinated retrieval and synthesis.  Specifically, we present the \textbf{SCOUT-RAG} framework, which operates in three coordinated stages and leverages four specialized agents. We first identify relevant knowledge domains, then perform adaptive local-to-global graph retrieval, and finally refine answers through iterative quality-guided evidence expansion. An illustrative overview of the framework is provided in Figure~\ref{fig:pipeline}.

\subsection{Stage I. Domain Relevance Assessment}

The first stage determines which distributed domains are likely to contain knowledge relevant to the query. Accurate relevance estimation is crucial: it suppresses unnecessary retrieval and prioritizes domains that meaningfully contribute to grounding, thereby improving cost-efficiency and cross-domain reasoning quality.

\paragraph{Domain Relevance Assessment Agent (DRAA).} Given distributed knowledge domains $\{\mathcal{D}_{i}\}_{i=1}^M$, we employ a \textbf{DRAA} to classify each domain into one of four relevance tiers: \texttt{HIGH}, \texttt{MODERATE}, \texttt{POTENTIAL}, and \texttt{IRRELEVANT}.

For each domain $\mathcal{D}_{i}$ and query $q$, three complementary information signals are computed:
\begin{itemize}
    \item \textbf{Semantic similarity:} 
    $s_i^{\text{sim}} = \mathrm{Sim}(q, \mathcal{D}_{i})$,  
    representing embedding-based cosine similarity between the query and domain-level representations.
    \item \textbf{Knowledge richness:} 
    $s_i^{\text{rich}} = |\mathcal{R}_i| / \max_j |\mathcal{R}_j|$,  
    quantifying relative data abundance by the normalized report count.
    \item \textbf{Historical performance:} 
    $s_i^{\text{hist}} = \frac{1}{|\mathcal{H}_i|} \sum_{h \in \mathcal{H}_i} Q(h)$,  
    capturing average answer quality over past queries associated with the domain.
\end{itemize}

These features jointly characterize domain–query compatibility from semantic, evidential, and behavioral perspectives. The DRAA combines them with query context to produce a relevance prediction:
\begin{equation}
    (\tau_i, r_i) = f_{\text{DRAA}}\!\left(q, \mathcal{D}_{i}, s_i^{\text{sim}}, s_i^{\text{rich}}, s_i^{\text{hist}}\right),
\end{equation}
where $\tau_i$ denotes the assigned relevance tier and $r_i$ is a natural-language rationale to support transparency and inspection. 

\subsection{Stage II. Domain-Scoped Seeding}

After Stage~I identifies candidate domains, Stage~II initializes retrieval and constructs the first round answer. This stage balances \emph{coverage} and \emph{precision}: high-confidence domains provide broad contextual grounding, while moderate-confidence domains contribute focused evidence. Two agents collaborate in this phase: the \textbf{Partial Answer Generation Agent (PAGA)} retrieves domain-specific evidence, and the \textbf{Overall Answer Synthesis Agent (OASA)} synthesizes these signals into an initial cross-domain response.

\paragraph{Partial Answer Generation Agent (PAGA). }
The PAGA executes domain-adaptive retrieval, guided by the relevance tiers $\tau_i$ from Stage~I. Retrieval is performed in parallel, but with differentiated granularity:
\begin{itemize}
    \item \texttt{HIGH}-relevance domains: \textbf{Global retrieval}, capturing broad semantic context via community-level summaries.
    \item \texttt{MODERATE}-relevance domains: \textbf{Local retrieval}, retrieving fine-grained entity-centric or relation-level evidence.
    \item \texttt{POTENTIAL} domains: Deferred,  held in reserve for possible activation in Stage~III.
    \item \texttt{IRRELEVANT} domains: Excluded to avoid unnecessary cost and noise.
\end{itemize}
Thus, for each domain $\mathcal{D}_i$ with graph $\mathcal{G}_i$, the partial answer is:
\begin{equation}
    \mathcal{A}_i =
    \begin{cases}
        f^{g}_{\text{PAGA}}(q, \mathcal{G}_i), & \text{if } \tau_i = \texttt{HIGH}, \\[4pt]
        f^{l}_{\text{PAGA}}(q, \mathcal{G}_i), & \text{if } \tau_i = \texttt{MODERATE}, \\[4pt]
        \emptyset, & \text{otherwise.}
    \end{cases}
\end{equation}
Here, $f^{g}_{\text{PAGA}}$ and $f^{l}_{\text{PAGA}}$ denote domain-specific global and local graph retrieval operators, respectively. This tiered policy ensures that retrieval cost concentrates only on domains with the highest expected payoff.

\paragraph{Overall Answer Synthesis Agent (OASA).} 
The OASA merges the partial answers into a coherent initial response:
\begin{equation}
    A^{(0)} = \Phi_{\text{OASA}}\!\left( \bigcup_{i=1}^{M} \mathcal{A}_i \right),
\end{equation}
where $\Phi_{\text{OASA}}(\cdot)$ performs cross-domain answer fusion with explicit source attribution and light consistency checking. The resulting $A^{(0)}$ serves as a grounded \emph{seed} answer: it establishes an initial scaffold, which Stage~III subsequently expands and refines as additional evidence is acquired.

\begin{figure}[t]
    \centering
    \includegraphics[width=\columnwidth]{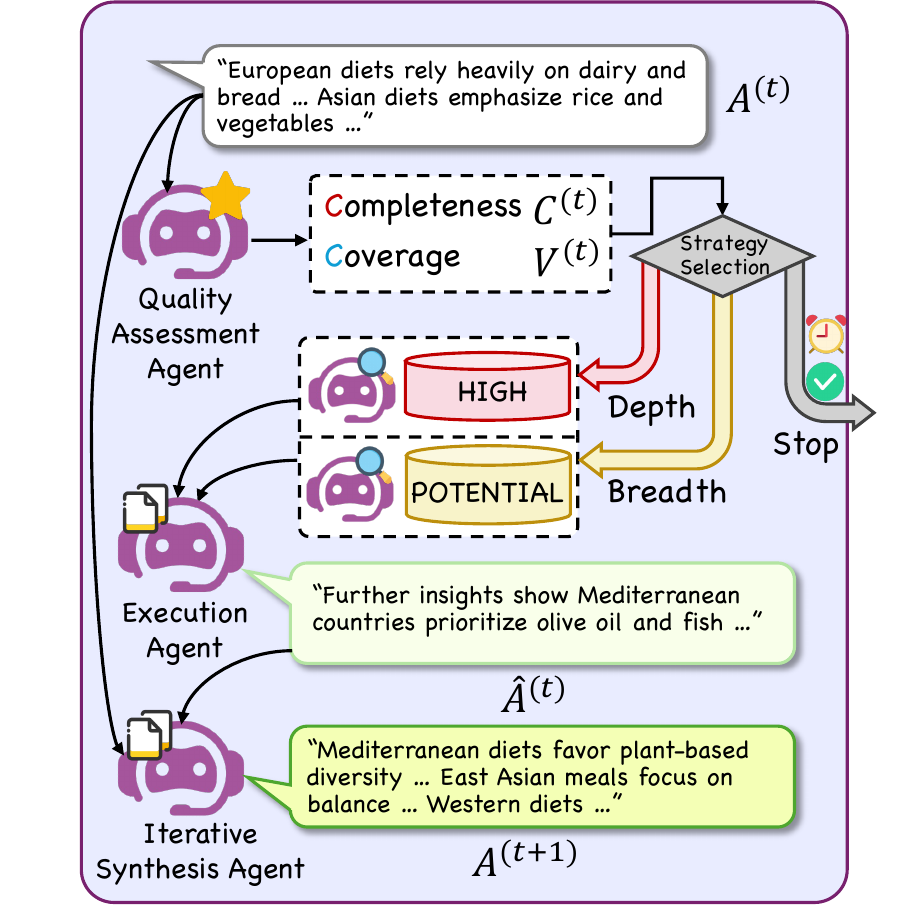}
    \caption{
        Illustration of Stage~III, where the system evaluates and improves answers through selective retrieval and synthesis. 
        \textit{Example query:} "What are the main differences in dietary habits among countries such as Japan, France, and the United States?".  
    }
    \label{fig:stage3}
\end{figure}

\subsection{Stage III. Iterative Cross-Domain Refinement}

Stage~III incrementally improves the initial seed answer from Stage~II through targeted, quality-guided refinement. The system evaluates the current answer, identifies missing content or uncertainty, and selectively activates retrieval to close knowledge gaps. This iterative loop continues until quality converges or the time budget is exhausted (Figure~\ref{fig:stage3}).

\paragraph{Answer Quality Assessment Agent (AQAA).}
The \textbf{AQAA} assesses the current answer $A^{(t)}$ using structured LLM-based evaluation prompts, producing a tuple of metrics and follow-up queries:
\begin{equation}
    (C^{(t)}, V^{(t)}, {G}^{(t)}, \mathcal{F}^{(t)}) = f_{\text{AQAA}}(A^{(t)}),
\end{equation}
where $C^{(t)}$ measures \textbf{completeness} (coverage of core conceptual elements),
$V^{(t)}$ measures \textbf{breadth} (coverage across relevant subtopics),
${G}^{(t)}$ denotes unresolved knowledge gaps,
and $\mathcal{F}^{(t)} = {q_k^f}$ is a set of targeted follow-up queries.
This structured evaluation provides actionable signals on what to refine next and where.

\paragraph{Strategy Selection.} 
A decision policy selects refinement mode $\alpha^{(t)}$ based on the current quality state and remaining time budget $T_{\mathrm{r}}$ according to our empirical evaluations:
\begin{equation}
\alpha^{(t)} = 
\begin{cases}
\text{Depth}, & \text{if } C^{(t)} < 0.75 \wedge T_{\text{r}} > 15\text{s}\\
\text{Breadth}, & \text{if } V^{(t)} < 0.70 \wedge T_{\text{r}} > 10\text{s}\\
\text{Hybrid}, & \text{if } C^{(t)} < 0.75 \wedge V^{(t)} < 0.70 \wedge T_{\text{r}} > 20\text{s}\\
\text{Stop}, & \text{otherwise}
\end{cases}
\label{statgyselection}
\end{equation}

Refinement terminates when:
\begin{itemize}
    \item Quality threshold met: $Q^{(t)} \ge 0.85$
    \item Time budget depleted: $T_{\text{r}} < 5\text{s}$
    \item Improvement stagnation: $\Delta Q^{(t)} = Q^{(t)} - Q^{(t-1)} < \epsilon$
\end{itemize}
where $Q^{(t)} = \frac{1}{2}(C^{(t)} + V^{(t)})$. 

Once a strategy is selected, refinement proceeds along one of three modes: a depth-focused pass that queries only \texttt{HIGH}-relevance domains using the targeted follow-up questions $\mathcal{F}^{(t)}$ to close core completeness gaps (similar in spirit to focused re-retrieval approaches such as DRIFT~\cite{edge2024local}); a breadth-expansion pass that activates previously dormant \texttt{POTENTIAL} domains to capture peripheral or emerging subtopics; or a hybrid mode that runs both paths in parallel when the answer is simultaneously incomplete and narrow. If none of the strategies are triggered, e.g., quality saturation or insufficient time, the system halts refinement.

\paragraph{Partial Answer Generation Agent (PAGA).} The \textbf{PAGA} executes the selected refinement mode by conducting targeted local retrieval:
\begin{equation}
    \hat{\mathcal{A}}^{(t)} = 
    \begin{cases}
        \bigcup_{i \in \mathcal{D}_{\text{high}}} f^{l}_{\text{PAGA}}(\mathcal{F}^{(t)}, \mathcal{G}_i), & \alpha^{(t)} = \text{Depth}, \\[4pt]
        \bigcup_{j \in \mathcal{D}_{\text{pot}}} f^{l}_{\text{PAGA}}(q, \mathcal{G}_j), & \alpha^{(t)} = \text{Breadth}, \\[4pt]
        \text{both above}, & \alpha^{(t)} = \text{Hybrid}.
    \end{cases}
    \label{eq:retrieval}
\end{equation}
Depth queries refine critical details in high-relevance domains, while breadth queries activate latent domains to capture overlooked context.

\paragraph{Overall Answer Synthesis Agent (OASA).}
The \textbf{OASA} integrates newly retrieved evidence with the current answer:
\begin{equation}
    A^{(t+1)} = \Phi_{\text{OASA}}\!\left(A^{(t)}, \hat{\mathcal{A}}^{(t)}\right).
\end{equation}
To ensure stability, the system tracks all intermediate candidates 
$\{A^{(0)}, A^{(1)}, \ldots, A^{(t)}\}$ 
and returns the highest-quality answer 
$A^{*}=\arg\max_i Qual(A^{(i)})$ upon termination,
ensuring resilience against late-stage drift or noisy retrieval. For clarity, we present a summary of SCOUT-RAG in Algorithm~\ref{alg:agentic-graphrag}.

\begin{algorithm}[t]
\caption{Scalable Cost-Efficient Cross-Domain Traversal for Distributed Agentic Graph-RAG}
\label{alg:agentic-graphrag}
\begin{algorithmic}[1]
\Require Query $q$, domains $\{\mathcal{D}_{i}\}_{i=1}^M$, time budget $T_{max}$
\Ensure Final answer $A^{*}$

\Statex \textbf{Stage 1: Domain Relevance Assessment}
\For{$D_i \in \mathcal{D}$}
    \State Compute domain relevance via DRAA:
    \Statex \hspace{1.5em}\texttt{HIGH}/\texttt{MODERATE}/\texttt{POTENTIAL}/\texttt{IRRELEVANT}
\EndFor

\Statex \textbf{Stage 2: Domain-Scoped Seeding}
\State Perform global retrieval on \texttt{HIGH} domains
\State Perform local retrieval on \texttt{MODERATE} domains
\State PAGA produces partial answers
\State OASA synthesizes seed answer $A^{(0)}$

\Statex \textbf{Stage 3: Iterative Cross-Domain Refinement}
\While{$T_r > 0$}
    \State Evaluate current answer $A^{(t)}$ via AQAA to obtain
    \State $(C^{(t)}, V^{(t)}, {G}^{(t)}, \mathcal{F}^{(t)})$
    \State Compute quality $Q^{(t)} = \frac{1}{2}(C^{(t)} + V^{(t)})$

    \State Select refinement strategy via Eq.~(\ref{statgyselection}):
    \If{$\alpha^{(t)}=\textsc{Stop}$}
        \State \textbf{break}
    \EndIf

    \State Conduct retrieval via PAGA according to $\alpha^{(t)}$
    \State Update answer via OASA fusion to obtain $A^{(t+1)}$
    \State Re-evaluate via AQAA
    \If{quality improves} update best answer $A^{*}$
    \EndIf
\EndWhile

\State \Return $A^{*}$

\end{algorithmic}
\end{algorithm}

\begin{table*}[t]
\centering
\caption{Average Performance Across Evaluation Dimensions. The best scores for centralized and decentralized methods are highlighted in bold.}
\renewcommand{\arraystretch}{0.93}
\setlength{\tabcolsep}{4pt}
\begin{tabular}{l|l|ccccc|cc}
\hline
\multirow{2}{*}{\textbf{Category}} & \multirow{2}{*}{\textbf{Method}} & \multicolumn{5}{c|}{\textbf{Quality}} & \multicolumn{2}{c}{\textbf{Cost}} \\ \cline{3-9}
 &  & \textbf{Comp.} ($\uparrow$) & \textbf{Div.} ($\uparrow$) & \textbf{Emp.} ($\uparrow$) & \textbf{Dir.} ($\uparrow$)& \textbf{Overall} ($\uparrow$) & \textbf{Time(s)} ($\downarrow$) & \textbf{Token} ($\downarrow$) \\ \hline
\multirow{3}{*}{\textbf{Centralized}} 
 & GraphRAG$_{\text{Local}}$ & 65 & 55 & 35 & 58 & 53 & \textbf{34.40} & \textbf{11,223} \\ 
 & GraphRAG$_{\text{Global}}$ & 60 & 50 & 30 & 55 & 49 & 45.89 & 640,574 \\ 
 & GraphRAG$_{\text{DRIFT-c}}$ & \textbf{72} & \textbf{70} & \textbf{45} & \textbf{63} & \textbf{63} & 231.85 & 693,731 \\ \hline
\multirow{2}{*}{\textbf{Decentralized}} 
 & GraphRAG$_{\text{DRIFT-dec}}$ & \textbf{90} & \textbf{88} & \textbf{75} & \textbf{88} & \textbf{85} & 414.88 & 879,911 \\ 
 & \textbf{SCOUT-RAG (Ours)} & 65 & 60 & 40 & 58 & 56 & \textbf{75.32} & \textbf{159,169}\\ 
 \hline
\end{tabular}
\label{tab:method_avg}
\end{table*}

\section{Experiments}
\subsection{Experimental Setup}
\subsubsection{Dataset}
To assess the scalability of our framework in handling a large number of domains, we employ Wikipedia articles from 45 countries as simulated domain datasets. Within each domain, a graph-based knowledge base is constructed, comprising 9–77 community reports. We also generate 100 queries with varying levels of domain coverage to comprehensively evaluate our framework, of which 89 are effectively answered by all methods. The queries are categorized into five levels: single-domain (20 queries, each involving one country), small multi-domain (20 queries, each involving five countries), medium multi-domain (20 queries, each involving ten countries), large multi-domain (20 queries, each involving twenty countries), and very large multi-domain (9 queries, each involving forty countries).

\subsubsection{Baseline}
We compare against four GraphRAG baselines across two deployment paradigms.
Under the centralized setting (all data from 45 countries consolidated into a single unified domain), we evaluate:
\begin{itemize}
    \item \textbf{GraphRAG$_{\text{Local}}$}: Local search for detailed entity-level information retrieval from specific community reports.
    \item \textbf{GraphRAG$_{\text{Global}}$}: Global search across hierarchical structures for comprehensive overviews via community summaries.
    \item \textbf{GraphRAG$_{\text{DRIFT-c}}$}: One global search followed by two refinement rounds (three follow-up local searches per round) to iteratively improve answer quality.
\end{itemize}
Under the decentralized setting (data distributed across 45 country-specific domains), we include:
\begin{itemize}
    \item \textbf{GraphRAG$_{\text{DRIFT-dec}}$}: Applies DRIFT independently in each domain—one global search plus two refinement rounds per domain, with final synthesis across all 45 domain-level answers.
\end{itemize}

\subsubsection{Mectric}
Following interactive agent evaluation methodologies~\cite{zhou2023sotopia}, we assess four dimensions adapted from GraphRAG~\cite{edge2024local}: Comprehensiveness (coverage of relevant aspects), Diversity (breadth of perspectives and sources), Empowerment (utility for follow-up exploration and actionable insights), and Directness (conciseness and clarity). Each dimension is scored 0–100 by GPT-4o as an LLM-as-judge using structured prompts.

All experiments were conducted on a system equipped with an NVIDIA RTX 3070 GPU with 9 GB VRAM (driver version 575.57.08). The compute environment consisted of 16 AMD EPYC 7B12 vCPUs, 31 GB memory.

\subsection{Main Results}

\subsubsection{Answer Quality Comparison} 
Table~\ref{tab:method_avg} summarizes the average performance across 89 queries spanning five complexity levels, evaluated by GPT-4o under two contrasting paradigms that comprise three centralized methods and two decentralized methods.


In decentralized mode, GraphRAG$_{\text{DRIFT-dec}}$ achieves the highest overall score (85) among all methods through exhaustive per-domain retrieval. Among centralized approaches, GraphRAG$_{\text{DRIFT-c}}$ reaches 63, outperforming GraphRAG$_{\text{Local}}$ (53) and GraphRAG$_{\text{Global}}$ (49) by more than 10 points. The proposed SCOUT-RAG system explores a third approach that combines decentralized deployment with intelligent and quality-controlled orchestration. It achieves an overall score of 56, which is 7 points lower than GraphRAG$_{\text{DRIFT-c}}$ but 7 points higher than GraphRAG$_{\text{Global}}$ and 3 points higher than GraphRAG$_{\text{Local}}$. This near-parity is notable given that SCOUT-RAG operates under distributed constraints without centralized coordination. The small performance gap indicates that quality-controlled orchestration effectively preserves answer quality within decentralized architectures.

Across individual evaluation dimensions, SCOUT-RAG demonstrates particularly strong performance in diversity (60), surpassing centralized baselines by 5–10 points. This improvement results from hierarchical domain categorization (\texttt{HIGH}/\texttt{MODERATE}/\texttt{POTENTIAL}) and quality-driven activation, which encourage systematic exploration of multiple perspectives. Although GraphRAG$_{\text{DRIFT-dec}}$ achieves the highest overall performance, it incurs an extreme computational cost of 414.88 seconds and 879,911 tokens, which is more than 5.5 times the resource consumption of SCOUT-RAG, as analyzed in the following section.

\subsubsection{Cost-Efficiency Analysis}
A critical advantage of SCOUT-RAG lies in resource efficiency. Our system consumes 159,169 tokens and completes queries in 75.32 seconds on average, representing 81.9\% token reduction and 81.8\% speedup compared to GraphRAG$_{\text{DRIFT-dec}}$ (879,911 tokens, 414.88s). Even against centralized GraphRAG$_{\text{Global}}$ (640,574 tokens), we achieve 75.2\% fewer tokens while maintaining comparable quality (56 vs. 49).

The cost and performance trade-off is particularly notable. GraphRAG$_{\text{DRIFT-dec}}$ attains the highest overall score (85) through exhaustive per-domain retrieval but requires 414.88s and 879,911 tokens, which is approximately 5.5 times SCOUT-RAG's execution time and token usage. Centralized GraphRAG$_{\text{DRIFT-c}}$ (63) also incurs significant overhead, consuming 231.85s and 693,731 tokens, which are 3.1 times and 4.4 times higher, respectively. Although GraphRAG$_{\text{Local}}$ executes faster (34.40s), it does so at the expense of diversity (55 vs. 60) and overall quality.

These efficiency gains enable practical deployment in resource-constrained scenarios. By achieving near-baseline performance with $>$80\% cost reduction compared to decentralized alternatives, SCOUT-RAG demonstrates the viability of quality-controlled orchestration for distributed knowledge retrieval at scale.

\begin{figure}[t]
    \centering
    \includegraphics[width=\columnwidth]{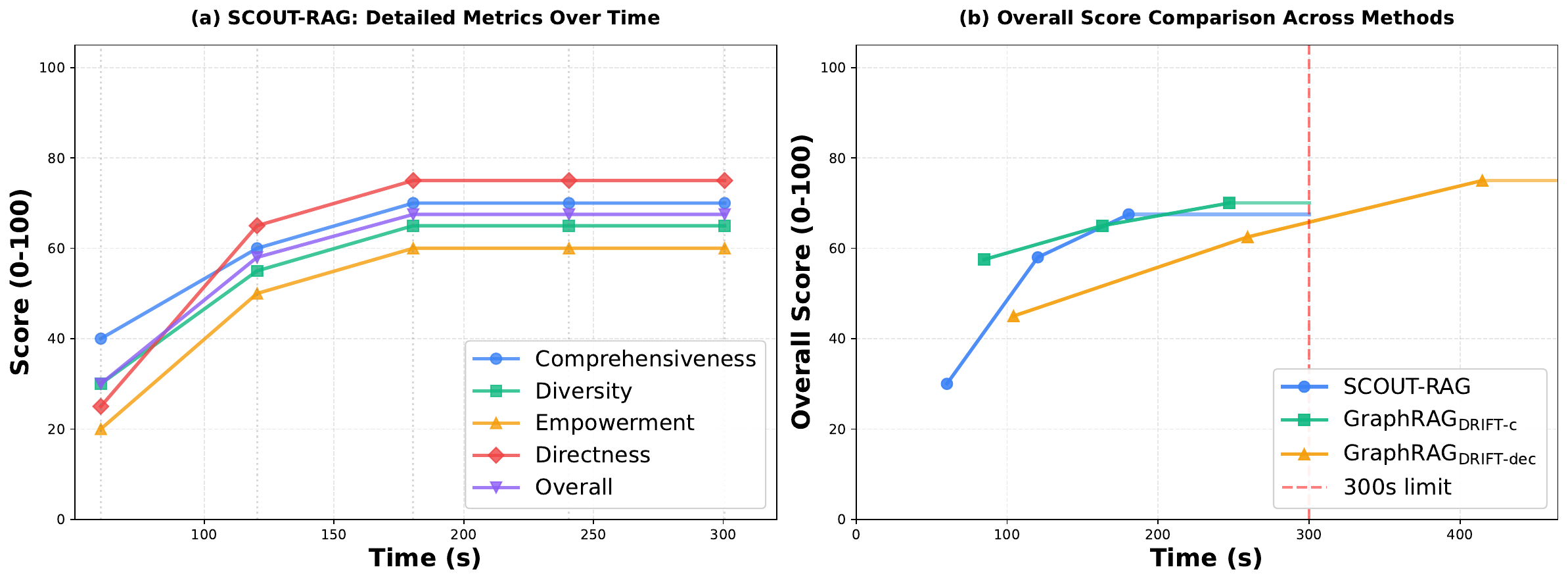}
    \caption{
        Time-performance monitoring and comparison.
    }
    \label{fig:methods_comparison}
\end{figure}

\begin{figure}[h!]
    \centering
    \includegraphics[width=0.95\columnwidth]{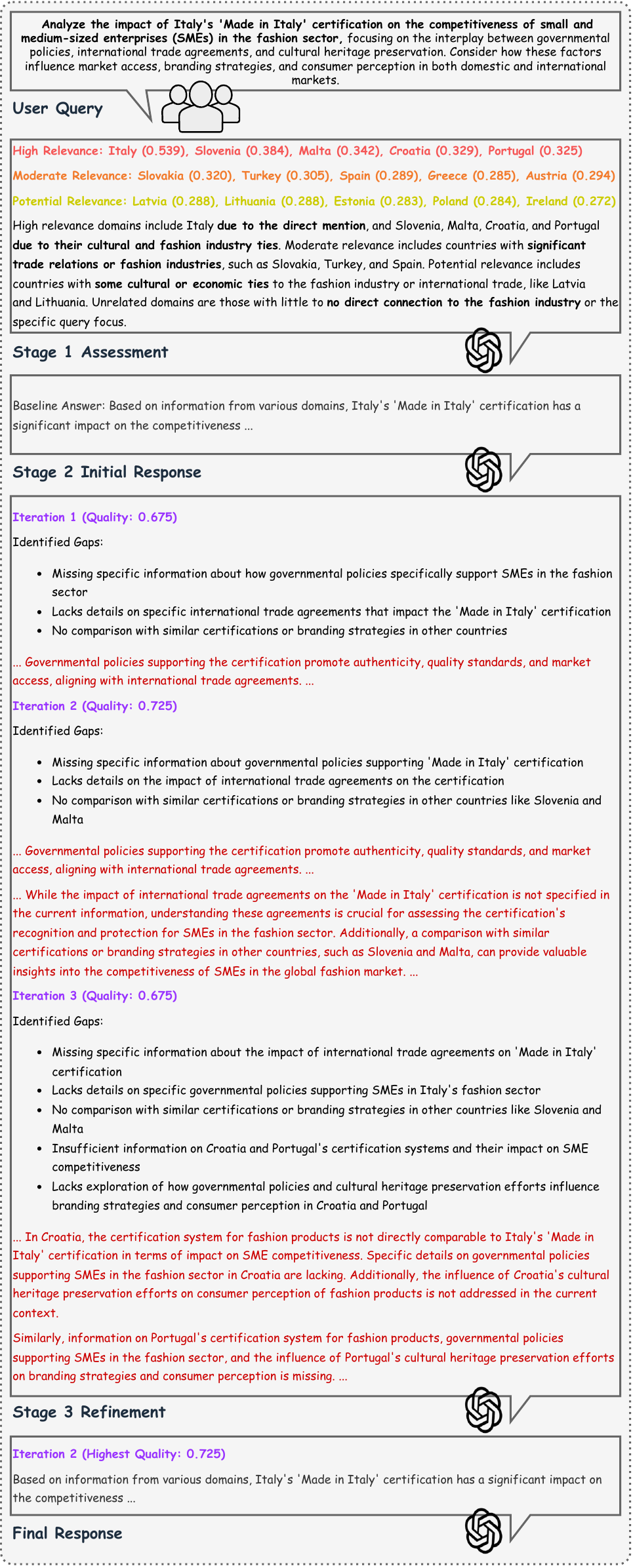}
    \caption{Case study on Italy's "\textit{Made in Italy}" certification.
    }
    \label{fig:casestudy}
\end{figure}

\subsection{A Closer Look at the Time Budget}

Figure~\ref{fig:methods_comparison} (left) demonstrates how SCOUT-RAG's answer quality evolves over time. All evaluation dimensions---Comprehensiveness, Diversity, Empowerment, and Directness---show rapid improvement in the first 120 seconds, followed by a performance plateau around 180s, indicating diminishing returns beyond this point. 
Figure~\ref{fig:methods_comparison} (right) compares three methods under different time constraints. SCOUT-RAG converges quickly to stable quality within the 300s budget. GraphRAG$_{\text{DRIFT-c}}$ achieves similar performance with comparable time efficiency. GraphRAG$_{\text{DRIFT-dec}}$, though requiring more time, ultimately reaches the highest overall score, demonstrating a trade-off between response speed and answer quality.

\subsection{Case Study}

We conducted a case study of SCOUT-RAG using a complex query. The overall workflow is presented in Figure~\ref{fig:casestudy}.

In Stage~I, the domain relevance module identified 10 relevant domains within 8.28 seconds. Italy ranked first (0.539), followed by Slovenia, Malta, Croatia, and Portugal, reflecting strong cultural and institutional similarity. This outcome demonstrates the system's ability to perform accurate geocultural alignment, ensuring that subsequent reasoning is grounded in comparable certification frameworks.

Stage~II produced an initial domain-scoped synthesis in 53.26 seconds. The system captured the main analytical dimensions concerning competitiveness, policy instruments, trade agreements, and heritage branding, but lacked concrete examples and cross-country references. AQAA rated breath at 0.65 and completeness at 0.70, prompting refinement to enhance specificity and comparative depth.

Stage~III executed three iterative refinement rounds over 41.12 seconds. The first round expanded retrieval to Slovenia, Malta, Croatia, and Portugal, introducing policy and branding details and improving coverage to 0.75. The second iteration focused on Italy, Malta, and Slovenia, achieving the highest overall quality (0.725) through contextual and evidential balance. The third round employed Breadth to explore \texttt{POTENTIAL} domains, but added minor noise and reduced final quality to 0.675. The best-track mechanism retained Iteration 2, preventing late-stage degradation.

    Overall, this case demonstrates that SCOUT-RAG effectively integrates policy, trade, and cultural heritage information across domains. Through adaptive retrieval, iterative synthesis, and best-track optimization, the system maintains analytical precision and interpretive coherence in addressing complex, interdisciplinary socio-economic questions such as "Made in Italy."

\section{Conclusion}
This work presented SCOUT-RAG, a scalable and cost-efficient agentic framework for retrieval-augmented generation across distributed knowledge domains. Unlike centralized Graph-RAG systems, SCOUT-RAG enables progressive cross-domain retrieval under explicit cost and privacy constraints. Empirical evaluations demonstrate that SCOUT-RAG achieves performance comparable to the centralized DRIFT, while maintaining a modest gap compared to the near-exhaustive Graph-RAG with DRIFT. Cost-efficiency analysis shows that SCOUT-RAG achieves significant reductions in both token consumption and execution time, demonstrating its potential for real-world deployments.

\section{Acknowledgments}
This research is supported by the National Research Foundation, Singapore under its National Large Language Models Funding Initiative (AISG Award No: AISG-NMLP-2024-003). Any opinions, findings and conclusions or recommendations expressed in this material are those of the author(s) and do not reflect the views of National Research Foundation, Singapore. This research is supported by A*STAR Career Development Fund $<$Project No. C243512010$>$.

\bibliography{aaai2026}

\end{document}